\documentclass[10pt, a4paper]{article}
\usepackage{lrec-coling2024} 

\usepackage{amsmath}
\usepackage{graphicx}
\usepackage{tabularx}
\usepackage{soul}
\usepackage{multirow}
\usepackage{arydshln}
\usepackage{array}
\usepackage{makecell}
\usepackage{contour}
\usepackage{amssymb}
\usepackage{color}
\usepackage{caption}
\usepackage{enumitem}

\title{A Challenge Dataset and Effective Models for Conversational Stance Detection}

\name{Fuqiang Niu$^{1,2}$, Min Yang$^{3}$, Ang Li$^{4}$\\
    {\bf \large Baoquan Zhang$^{4}$, Xiaojiang Peng$^{2}$, Bowen Zhang$^{2*}$\thanks{* Corresponding author.}}
    }
    
\address{$^{1}$Shenzhen University, China \ \ \ \  $^{2}$Shenzhen Technology University, China \\
        $^{3}$Shenzhen Institute of Advanced Technology, Chinese Academy of Sciences, China\\
        $^{4}$Harbin Institute of Technology, Shenzhen, China\\
         nfq729@gmail.com \ \ \ \ min.yang@siat.ac.cn\\
         angli@stu.hit.edu.cn \ \ \ \  baoquanzhang@yeah.net\\
         pengxiaojiang@sztu.edu.cn \ \ \ \ zhang\_bo\_wen@foxmail.com \\}

\abstract{
Previous stance detection studies typically concentrate on evaluating stances within individual instances, thereby exhibiting limitations in effectively modeling multi-party discussions concerning the same specific topic, as naturally transpire in authentic social media interactions. This constraint arises primarily due to the scarcity of datasets that authentically replicate real social media contexts, hindering the research progress of conversational stance detection. In this paper, we introduce a new multi-turn conversation stance detection dataset (called \textbf{MT-CSD}), which encompasses multiple targets for conversational stance detection. To derive stances from this challenging dataset, we propose a global-local attention network (\textbf{GLAN}) to address both long and short-range dependencies inherent in conversational data. Notably, even state-of-the-art stance detection methods, exemplified by GLAN, exhibit an accuracy of only 50.47\%, highlighting the persistent challenges in conversational stance detection. Furthermore, our MT-CSD dataset serves as a valuable resource to catalyze advancements in cross-domain stance detection, where a classifier is adapted from a different yet related target. We believe that MT-CSD will contribute to advancing real-world applications of stance detection research. 
Our source code, data, and models are available at \url{https://github.com/nfq729/MT-CSD}.
 \\ \newline  }
\begin{document}

\maketitleabstract

\section{Introduction}
In contemporary social media platforms, users frequently express their viewpoints on contentious subjects related to specific targets. The aggregation and analysis of these expressed perspectives can unveil prevailing trends and opinions concerning controversial topics, ranging from issues like abortion to epidemic prevention~\cite{glandt2021stance}. This wealth of information holds significant promise for applications in web mining and content analysis. The insights derived from such analyses can serve as a valuable resource for various decision-making processes, including but not limited to advertising recommendations and presidential elections~\cite{li2021p, zhang2023twitter}. Consequently, automatic stance detection on social media has emerged as a pivotal approach within the domain of opinion mining, facilitating a deeper understanding of user opinions on diverse issues.

\begin{figure}
\centering
\includegraphics[width=\linewidth]{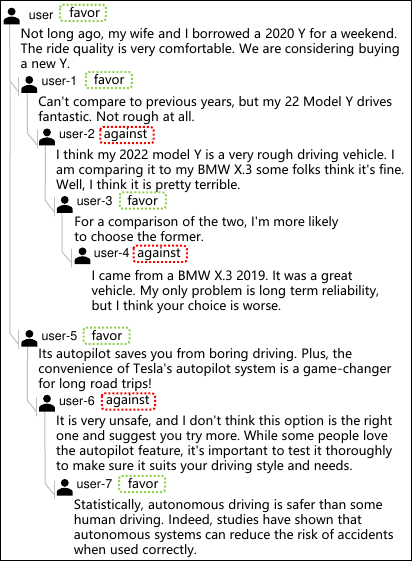}
\caption{An example of conversational stance detection.} 
\label{fig:example}
\end{figure}

Stance detection endeavors to classify the polarity of attitudes expressed in textual content (e.g., statements, tweets, articles, or comments) towards a specific target~\cite{MohammadKSZC16}. Existing studies are typically categorized into target-specific, cross-target, and zero-shot stance detection, with a predominant focus on analyzing individual sentences~\cite{AllawayM20}. However, in the context of social media analysis, users commonly articulate their perspectives through conversational exchanges. Conventional context-free stance detection methods encounter challenges in accurately predicting stances in such conversational settings. For instance, Figure~\ref{fig:example} illustrates a social media discussion. Within this conversational thread, it is difficult to detect the stances of $user_{3}$ and $user_{4}$ towards Tesla without the contextual backdrop of preceding interactions. In addition, following $user_{5}$'s input, the discussion diversifies into various Tesla-related topics, such as ``autopilot'', providing valuable cues for discerning stances in subsequent comments. Consequently, conversational stance detection (CSD), which aims to identify stances within conversation threads, has garnered increased attention in recent research.

To date, two CSD datasets have been developed and served as benchmarks for CSD tasks, namely SRQ~\cite{villa2020stance} and Cantonese-CSD (CANT-CSD)~\cite{li2022improved}. However, these two datasets have several limitations: (i) the existing datasets predominantly feature examples with few reply turns. For instance, the SRQ dataset comprises solely direct reply data, representing 1-turn comments. Similarly, in the CANT-CSD dataset, merely 6.3\% of the data encompasses more than 3 reply turns; (ii) the annotation quality in existing datasets falls short of optimal standards. Notably, the SRQ dataset only annotates the stance of the reply text, neglecting to annotate the original comment; (iii) the CANT-CSD dataset is exclusively in Cantonese and suffers from a scarcity of labeled examples. These issues limit the application of CSD models in real social media scenarios. Therefore, constructing a high-quality CSD dataset is essential.


To foster advancements in CSD research, we introduce a multi-turn conversation stance detection dataset (called \textbf{MT-CSD}), which encompasses 15,876 meticulously annotated instances, representing a substantial increase in scale compared to previous stance detection datasets. 
A noteworthy characteristic is the high prevalence of comments with a depth exceeding 4 turns, constituting 75.99\% of the dataset. In particular, in contrast to CANT-CSD, the only other dataset featuring multi-turn (more than two-turn) comments, MT-CSD exhibits over 12 times more instances with a depth of 4, providing a more extensive and diverse set of conversational data for stance modeling. The MT-CSD dataset introduces distinctive challenges for stance detection: (i) implicit target references embedded within local sub-discussions necessitate a nuanced understanding of contextual information; and (ii) while the posts directly mentioning targets offer explicit stance cues, determining stance in comments demands a more intricate process involving the resolution of coreferences and reliance on other contextual clues. 

To tackle the aforementioned challenges, we introduce a novel global-local attention network (GLAN) designed specifically for CSD. The GLAN architecture adopts a three-branch structure to address the intricacies of conversational dynamics comprehensively. The first branch incorporates a global attention network aimed at capturing long-range dependencies. The second branch utilizes convolutional neural networks to detect subtle, local conversational nuances by focusing on smaller segments of dialogue, thereby providing a granular analysis of the discourse. The third branch leverages graph convolutional networks to capture nuanced local discussion segments within the broader conversation.


The main contributions of this paper can be summarized as follows:
\begin{itemize}[leftmargin=*]
\item  We introduce a challenging multi-turn conversation stance detection (MT-CSD) dataset tailored for conversational stance detection. This dataset is the largest human-labeled English conversational stance dataset to date. The release of MT-CSD would push forward the research of CSD.
\item We propose a novel GLAN architecture featuring an upper global branch that learns from the reply dependency graph and a lower local branch that captures discussion segments within the conversation.
\item We conduct a comprehensive performance evaluation of state-of-the-art stance detection methods employing three widely adopted methodologies: fine-tuning with deep neural networks, prompt-tuning with pre-trained language models (PLMs), and in-context learning with large language models (LLMs). Experimental findings shed light on the challenges faced by current models in CSD. 
\end{itemize}

\section{Related Work}

\subsection{Stance Detection Datasets}
To date, several datasets have been curated and have emerged as benchmark datasets for stance detection in the realm of social media. The characteristics of these datasets are presented in Table ~\ref{tab:datasets}. SemEval-2016 Task 6 (SEM16) stands as the inaugural stance detection dataset sourced from Twitter and holds prominence as a widely used benchmark, comprising 4,870 tweets expressing stances towards various targets \cite{MohammadKSZC16}. Subsequently, to leverage large-scale annotated datasets, \citet{zhang2020enhancing} extended SEM16 by introducing the \textit{Trade Policy} target. \citet{conforti2020will} contributed to the WT-WT dataset encompassing a more extensive labeled corpus. Additionally,  \citet{li2021p} introduced the P-Stance dataset, specifically tailored to the political domain, featuring tweets with a longer average length. \citet{glandt2021stance} presented a dataset designed for COVID-19-Stance detection. In complement to the aforementioned stance detection datasets, designed for specific targets, the VAST dataset was proposed by \citet{allaway2020zero}, focusing on zero-shot stance detection with a diverse array of over a thousand targets. Notably, these efforts primarily center around sentence-level (individual post-level) stance detection tasks. 

Currently, there exist only two CSD datasets specifically tailored for comments within conversation threads. The SRQ dataset \cite{villa2020stance} is introduced to address stance detection within tweet replies and quotes. However, the SRQ dataset concentrates solely on single-turn replies and quotes. The CANT-CSD dataset \cite{li2022improved} is designed to address stance detection in multi-turn conversation scenarios. Despite its comprehensive coverage, most data in CANT-CSD is confined to shallow reply rounds. Specifically, 80.1\% of the data comprises two rounds of replies, with only 6.3\% featuring more than three rounds. Our observations indicate that, particularly under the influence of trending topics, the depth of comment replies can frequently surpass five rounds, thereby constraining the applicability of CANT-CSD in real-world scenarios. Furthermore, the CANT-CSD dataset is annotated in Cantonese, limiting its broader impact within the stance detection community.

\begin{table}
  \resizebox{\linewidth}{!}{
  \begin{tabular}{c|c|c|c|c}
    \hline
    \textbf{Type} & \textbf{Sentence-level} & \multicolumn{3}{c}{\textbf{Conversation-based}}\\
    \hline
    \textbf{Classif. Task} &  Sentence classification & \multicolumn{3}{c}{\makecell[c]{ Conversation history \\classification}} \\
    \hline
    \textbf{Label} & \multicolumn{4}{c}{Favor, Against, None} \\
    \hline
    \textbf{Work} & \makecell[c]{SEM16, P-stance\\ COVID-19-Stance\\ VAST, WT-WT} & SRQ & CANT-CSD & \textbf{\makecell[c]{Our\\work}}\\
    \hline
    \textbf{Target-nums} & $\geq $4  & 4 & 1 & 5\\
    \hline
    \textbf{Multi-turn} & - & $\times$ & $\checkmark$ & $\checkmark$\\
    \hline
   
 \textbf{English}& $\checkmark$& $\checkmark$& $\times$&$\checkmark$\\
  \hline
  \end{tabular}
  }
\caption{\label{tab:datasets} Comparison of different stance detection datasets.}
\end{table}

\subsection{Stance Detection Approaches}
The objective of stance detection is to discern the expressed attitude of a given text towards a specific target \cite{JainJDS22, RaniK22, li-etal-2023-stance}. Conventional approaches in this domain predominantly pertain to sentence-level stance detection, categorized into in-target, cross-target, and zero-shot stance detection. 

In the in-target setup, conventional methods often leverage deep neural networks, such as attention networks and GCN, to train a stance classifier. The attention-based methods utilize target-specific information as the attention query, deploying an attention mechanism to infer the stance polarity \cite{dey2018topical, wei2018multi, du2017stance, sun2018stance}. The GCN-based methods utilize GCN to model the relation between target and input text \cite{LiPLSLWYH22,CignarellaBR22,ConfortiBPGTC21}.

Cross-target stance detection (CTSD) tasks have garnered attention in various studies, which can be classified into two categories. The first category employs word-level transfer methods, utilizing common words shared by two targets to bridge the knowledge gap \cite{augenstein2016stance}. The second category tackles the cross-target problem by leveraging concept-level knowledge shared by two targets \cite{wei2019modeling, zhang2020enhancing, cambria2018senticnet,ding2024cross}.

Zero-shot stance detection (ZSSD) involves unseen targets for a trained stance detection model, presenting a more challenging task. \citet{AllawayM20} introduced a large-scale human-labeled stance detection dataset designed for zero-shot scenarios. \citet{AllawayM20} utilized a target-specific stance detection dataset for ZSSD, employing adversarial learning to extract target-invariance information. \citet{liu2021enhancing} proposed a common sense knowledge-enhanced graph model based on BERT, leveraging inter- and extra-semantic information. Additionally, \citet{liang2022zero} presented an effective method to distinguish target-invariance from target-specific features, facilitating a more robust learning of transferable stance features.

\section{Dataset Construction}
In this section, we provide a comprehensive exposition of the creation process and unique attributes of our MT-CSD dataset comprising 15,876 texts sourced from Reddit.

\subsection{Data Collection}
To procure authentic social media interaction data, we leveraged Reddit, renowned as one of the largest and most extensive forums, to ensure the richness and authenticity of the collected CSD data. We accessed the data from Reddit through the official API provided by the platform\footnote{https://www.reddit.com/dev/api}. During the data collection process, we collected Reddit posts and associated popularity metrics such as upvotes and comment counts. A manual review of the posts was conducted to assess their relevance to the given targets, guaranteeing that the collected posts were highly pertinent and featured sufficiently in-depth comments to support dataset annotation. Then, we collected comments for each selected post. The resulting dataset encompassed relevant posts, associated discussions, and comments, providing a comprehensive overview of conversations centered around the specified targets. The selected targets for this dataset included ``\textit{Tesla}'', ``\textit{SpaceX}'', ``\textit{Donald Trump}'', ``\textit{Joe Biden}'', and ``\textit{Bitcoin}''. 

\subsection{Data Preprocessing}
To ensure the high quality of this MT-CSD dataset, we implemented several rigorous preprocessing steps:
\begin{itemize}[leftmargin=*]
    \item High Relevance to Target: The content of each post has to be highly relevant to the specified target. A two-reviewer process was employed to assess such relevance, with only posts deemed highly relevant by both reviewers retained.
    \item Minimum 200 Comments per Post: To ensure each post garnered significant attention and discussion, we set a requirement of at least 200 comments per post. Insufficient comment counts would result in inadequate conversation depth and reduced complexity.
    \item Appropriate Text Length: Constraints were imposed on the text length of posts. To ensure data quality, the post length had to be at least 15 words but no more than 150 words. Texts with less than 15 words are either too simplistic for detecting stance or too noisy, while posts with more than 150 words often contain duplicate expressions.
    \item Excluding Non-English Posts: As we aim to construct an all-English dataset, non-English language posts were systematically removed to maintain language consistency. Multilingual stance detection is left as a potential avenue for future exploration.
\end{itemize}
Following this stringent data filtering process, the resulting data distribution is summarized in Table~\ref{tab:post}.
 \begin{table}
  
    \resizebox{\linewidth}{!}{
  \begin{tabular}{cccccc}
    \hline
    \textbf{Target} & \textbf{Bitcoin} & \textbf{Tesla} & \textbf{SapceX} & \textbf{Biden} & \textbf{Trump}  \\
    \hline
    \textbf{Post} & 93 & 52 & 32 & 72 & 81  \\
    \textbf{Comment} & 9,716 & 8,989 & 4,911 & 10,593 & 10,203  \\
    \hline
  \end{tabular}
  }
\caption{\label{tab:post} The number of data items for each target.}
\end{table}

\subsection{Data Annotation and Quality Assurance}
We implemented an annotation system to meticulously ensure that annotators rigorously reviewed the preceding context and provided accurate attitude labels. This system is tailored to conversational data and aims to streamline and enhance the process of comprehensive data annotation. During the annotation process, explicit guidelines were provided to annotators, instructing them to label each comment with ``\textit{against}'', ``\textit{favor}'', or ``\textit{none}'' to indicate their attitude. Additionally, annotators were prompted to specify whether newly added comments were related to the specified target.


We invited eleven researchers possessing expertise in natural language processing (NLP) to annotate the data. Prior to the formal annotation process, we adopted two pilot annotation rounds to ensure the reliability of the annotated data. Three additional expert annotators reviewed the pilot annotated data to ensure each annotator could effectively perform the annotation task. In the formal annotation stage, we ensured that each data instance was annotated by at least two annotators. When there was disagreement between the two initial annotators, an additional annotator  were involved in labeling the contentious statements, and a final consensus was reached through voting. This annotation approach not only ensured the reliability of the data, but also integrated inputs and consensus from multiple annotators, improving the overall quality of stance labels assigned to each instance.
After obtaining the annotation results, we computed the kappa statistic~\cite{kappa} and inter-annotator agreement as measures of inter-annotator agreement. Following \cite{li2021p}, we selected the ``\textit{Favor}'' and ``\textit{Against}'' classes to compute the kappa statistic values. The results are presented in Table~\ref{tab:consistency}. The results indicate that the kappa statistic for all five targets exceeds 70\%, with an average score of 83\%. The average score for inter-rater consistency among multiple annotators, where agreement was rated as one and disagreement as 0, is 76\%, affirming that our dataset is well-annotated and of high quality.


\begin{table}
  
    \resizebox{\linewidth}{!}{
  \begin{tabular}{ccccccc}
    \hline
    \textbf{Target} & \textbf{Bitcoin} & \textbf{Tesla} & \textbf{SapceX} & \textbf{Biden} & \textbf{Trump}&\textbf{Avg.} \\
    \hline
    \textbf{consistency} & 0.79& 0.75 & 0.79 & 0.71& 0.74&0.76 \\
    \textbf{kappa} & 0.93& 0.74 & 0.83 & 0.96 & 0.71&0.83 \\
    \hline
  \end{tabular}
  }
  \caption{\label{tab:consistency} Annotation consistency and agreement.}
\end{table}

\begin{table}
  
    \resizebox{\linewidth}{!}{
  \begin{tabular}{cccc}
    \hline
    \textbf{Instance} & \textbf{Avg. WC} & \textbf{Depth} & \textbf{Number}\\
    \hline
    
    \textbf{Post} & 18.02& 1 & 218 (1.37\%) \\
    \hline
    \multirow{7}{*}{\textbf{Comment}}
     & 26.48& 2 & 1,017 (6.41\%)\\
     & 29.09& 3 & 2,575 (16.22\%)\\
     & 31.50& 4 & 3,250 (20.47\%)\\
     & 31.97& 5 & 3,204 (20.18\%)\\
     & 33.62& 6 & 2,739 (17.25\%)\\
     & 35.44& 7 & 1,900 (11.97\%)\\
     & 38.33& 8 & 973 (6.12\%)\\
    \hline
    
  \end{tabular}

  }
  \caption{\label{tab:statistics} Statistics of the MT-CSD dataset. Here, WC is short for word count.}
\end{table}
\begin{table*}
\begin{center}	
  \resizebox{\linewidth}{!}{
\begin{tabular}{c|ccc|ccc|ccc|ccc|ccc}
\hline
\textbf{Target} & \multicolumn{3}{c|}{\textbf{Bitcoin}} & \multicolumn{3}{c|}{\textbf{Tesla}} & \multicolumn{3}{c|}{\textbf{SpaceX}} & \multicolumn{3}{c|}{\textbf{Biden}} & \multicolumn{3}{c}{\textbf{Trump}} \\  
\hline
\textbf{depth} &against  & favor  & none & against  & favor  & none& against  & favor  & none& against  & favor  & none& against  & favor  & none  \\ 

\hline
    \textbf{1-2}       & 100   & 92 & 110      & 110 & 13 & 147      & 23 & 81 & 84     &9  &82 &85    &135 &11 & 153\\
    \textbf{3-5}       & 791  & 561 &689    & 638  & 235 & 1116   & 191 & 356 & 687  &194 &605 &818    &1006  &120  & 1023\\
    \textbf{6-8}       & 433  & 216 & 393   & 398 & 229& 805      & 84 & 158 & 391   &149 &499 &506    &526 &76 & 748\\
\hline
    \textbf{class-all}       & 1324  & 869 & 1192   & 1146 & 477& 2068      & 298 & 595 & 1162   &352 &1186 &1409
  &1667 &207 & 1924\\
\hline
    
\textbf{all} & \multicolumn{3}{c|}{3385} & \multicolumn{3}{c|}{3691} & \multicolumn{3}{c|}{2055} & \multicolumn{3}{c|}{2947} & \multicolumn{3}{c}{3798} \\ 
\hline
\end{tabular}
}
\end{center}	
\caption{\label{tab:Data distribution} Statistics of the MT-CSD dataset with varying input depths.}
\end{table*}
\subsection{Data Analysis}
Table \ref{tab:Data distribution} presents the statistics of our MT-CSD dataset. The final annotated dataset comprises 15,876 instances, which is 2.7 times and 3.4 times larger than the CANT-CSD and SRQ datasets, respectively. Table \ref{tab:statistics} provides the distribution of instances across different depths. A significant portion, 75.99\%, of the data in our MT-CSD dataset has a depth greater than 3. In comparison, only 6.3\% of the CANT-CSD dataset exceeds depth 3. We create training and testing sets for all targets in an 80/20 ratio. During experiments, we randomly select 15\% of the data from the training set as a validation set. 


\subsection{Challenges}
Our MT-CSD dataset is a challenging dataset for several reasons:
\begin{itemize}[leftmargin=*]
    \item Implicit target references: In MT-CSD, targets are referenced more implicitly. For instance, as illustrated in Figure~\ref{fig:example}, discussions about the given target ``\textit{Tesla}'' expand to include discussions of ``\textit{autopilot}'' as comment depth increases. In essence, the stance towards the target is expressed more implicitly in local discussions within the full conversation. This complexity demands effective recognition and understanding of these local discussion segments to identify stances correctly.
    \item Coreference relations: Generally, posts explicitly mention the target and contain richer stance-bearing words, making it relatively easier to discern the stance towards the target. Different from posts used in most previous datasets, comments often exhibit contextual dependencies such as coreference relations, introducing challenges for stance detection models.
\end{itemize}

\section{Our Methodology}
\begin{figure}
\centering
\includegraphics[width=\linewidth]{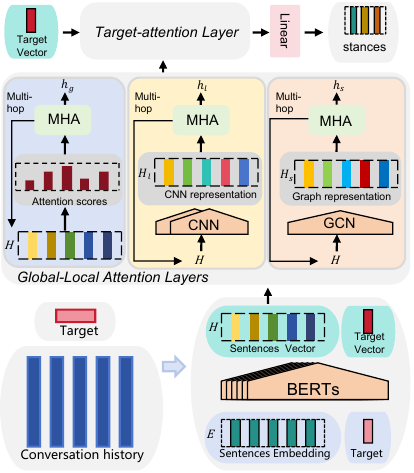}
\caption{The architecture of our GLAN framework.} 
\label{fig:model}
\end{figure}

In this section, we present a detailed description of our proposed global-local attention network (GLAN) model for conversational stance detection. As illustrated in Figure~\ref{fig:model}, the GLAN model comprises three key modules: the text representation layer, the global-local attention layers, and the target-attention layer. The text representation layer utilizes BERT as the backbone to generate a contextualized representation for each token in the input conversation text. The global-local attention layer includes three integral parts: the global part, the local part, and the structural part. The target-attention layer operates on the vector derived from the global-local attention layer and performs an attention operation on the target, producing the final result.

\subsection{Text Representation Layer}

We utilize BERT to generate deep contextualized representations for input conversation. Specifically, we represent the conversation $X = \langle x_1, x_2, \ldots, x_n \rangle$ as a sequence of $n$ utterances, where each utterance $x_i = \langle w_{i,1}, w_{i,2}\ldots, w_{i,j} \rangle \quad (\forall j = 1, \ldots, l_i)$ represents a post or a comment. To extract contextual information from utterances $\langle x_1, x_2, \ldots, x_{n-1} \rangle$, we concatenate all instances in $X$ into a token sequence $S_x$ in which every two consecutive instances are separated with a special token $[SEP]$. Subsequently, we utilize a BERT tokenizer to transform $S_x$ into BERT's input embeddings $E$. Subsequently, we derive a vector representation for each sentence, denoted as $h_{x_i}$, by taking the average of the constituent word vectors. Finally, we obtain the sentence vectors for the entire conversation $H$ by combining and aggregating the individual sentence vectors $h_{x_i}$.

\subsection{The Global-Local Attention Layer}
After obtaining sentence embeddings generated by a pre-trained BERT model, our approach involves operations at three distinct modules. In each module, unique sentence embedding vectors are obtained, and a common operation denoted as multi-hop attention (MHA) is applied. First, we present an overview of the operations for obtaining various sentence embedding vectors from each module. Subsequently, we provide a description of the shared MHA operation.


\paragraph{Global Layer}
To capture long-range dependencies between the text and its dialogue history, we develop a global layer. Initially, we perform a multiplication operation involving the last sentence vector, denoted as $h_{x_n}$, and all the preceding sentence vectors. Subsequently, we apply the softmax activation function to obtain an attention scores matrix, denoted as $\gamma$:
\begin{equation}
\gamma_t = softmax(h_{x_n}^T h_t) \label{eq:weights}
\end{equation}
where $h_t$ denotes the \textit{t}-th sentence vector from $H$.
We conduct a multiplication operation involving the weight matrix and the feature vector, resulting in the generation of a new matrix of sentence vectors denoted as:
\begin{equation}
H_g =  \sum_{t=1}^n \gamma_t h_t \label{eq:weights_2}
\end{equation}

\paragraph{Local Layer}

The acquired sentence embedding vectors undergo processing through two one-dimensional convolutional layers with a kernel size of 2, yielding the modified sentence embedding vectors referred to as $H_l$. These vectors maintain the same dimensionality as the original sentence vectors but are enriched with localized information.

\paragraph{Structural Layer}
Subsequently, we propose a structural layer that enables the model to leverage comment relations for sentence representation generation. First, we construct a comment graph (CG) from the conversation history, where nodes represent sentence vectors $h_{x_i}$ and edges denote comment relations. We represent the adjacency matrix of CG as $A$. After obtaining the $H$, we feed them into a two-layer GCN. The graph representation $H_s$ can be calculated as:
\begin{equation}
\label{1}
H_s =\sigma(A \sigma ({A}HW_0)W_1)
\end{equation}
where $\sigma$ represents a non-linear function, and $W_0$ and $W_1$ are trainable parameters.

\paragraph{MHA}
After obtaining the sentence vectors ($H_g$, $H_l$, $H_s$) from the three distinct modules, they are subsequently utilized as input for the MHA module.

The MHA module follows a methodology akin to the MemN2N~\cite{NIPS2015_8fb21ee7} module. 
Initially, it undergoes an attention operation ($Att$), mirroring the process in the global layer. 
Then, it proceeds through an activation function and layer normalization, following which it is subjected to multiplication by the variable $\lambda$ and addition to the original sentence embedding vectors.
\begin{gather}
    H^2_g = \lambda LN(\sigma(Att(H_g))) + H\\
    H^2_l = \lambda LN(\sigma(Att(H_l))) + H\\
    H^2_s = \lambda LN(\sigma(Att(H_s))) + H
\end{gather}
where $LN$ represents layer normalization, $\sigma$ represents the sigmoid activation function. we repeat MHA module three times to obtain sentence vectors. Finally, we sum up the obtained sentence vectors, resulting in a vector of dimension $\mathbb{R}^{1 \times h}$. We represent the summed sentence vectors as $h_g$, $h_l$ and $h_s$, respectively.

\subsection{The Target-attention Layer}
In the target-attention layer, we employ the target vector derived from the pre-trained BERT model as a query and execute an attention operation with the resulting sentence vectors (e.g., $h_g$, $h_l$, $h_s$) as depicted in the diagram. Finally, we concatenate the obtained vectors and pass them through a fully connected layer to obtain the stance.

Given an annotated training set, we utilize the cross-entropy between the predicted stance and the ground-truth stance as our loss function for stance detection.





\section{Experimental Setup}
In this section, we present the evaluation metrics utilized in the experiments and outline the baseline methods employed for the evaluations.

\subsection{Evaluation Metrics}
We adopt $F_{avg}$ as the evaluation metric to evaluate the performance of stance detection methods, similar to~\cite{li2021p} and ~\cite{10.1145/3003433}. $F_{avg}$ represents the average F1 score computed for the ``against'' and ``favor'' stances. We compute the $F_{avg}$ for each target. 

\subsection{Baseline Methods}
We conduct extensive experiments with state-of-the-art stance detection methods, which can be divided into four categories: supervised training with DNNs, prompt-tuning with PLMs, fine-tuning with PLMs, and in-context learning with LLMs. 

\paragraph{Supervised Training with DNNs}
We adopt several widely-used DNNs as baselines: (i) \textbf{BiLSTM}~\cite{650093} is trained to predict the stance towards a target without explicitly using target information; (ii) \textbf{GCAE}~\cite{xue-li-2018-aspect} is a CNN model that utilizes a gating mechanism to block target-unrelated information; (iii) \textbf{TAN}~\cite{ijcai2017p557} is an attention-based BiLSTM model; and (iv) \textbf{CrossNet}~\cite{ijcai2017p557} adds an aspect-specific attention layer before classification. 

\paragraph{Prompt-tuning with PLMs}
Three representative prompt-tuning methods with PLMs are compared: (i) \textbf{MPT}~\cite{KEPrompt} develops prompt-tuning-based PLM to perform stance detection, where humans define the verbalizer; (ii) \textbf{KPT}~\cite{shin2020autoprompt} introduces external lexicons to define the verbalizer. Different from the lexicon utilized in Reference~\cite{shin2020autoprompt}, KPT utilize SenticNet instead of sentiment lexicons; and (iii) \textbf{KEPrompt}~\cite{KEPrompt} uses an automatic verbalizer to automatically define the label words. All three Prompt-tuning with PLMs are based on bert-base-uncased\footnote{https://huggingface.co/bert-base-uncased}.

\paragraph{Fine-tuning with PLMs}

Four representative methods performing fine-tuning with PLMs are employed as baselines: (i) the pre-trained \textbf{BERT}~\cite{devlin-etal-2019-bert} is fine-tuned on the training data; (ii) \textbf{JoinCL}~\cite{liang2022jointcl} employs stance contrastive learning and target-aware prototypical graph contrastive learning for stance detection, which are expected to generalize target-based stance features to unseen targets; (iii) \textbf{TTS}~\cite{TTS} utilizes target-based data augmentation to extract informative targets from each training sample and then utilizes the augmented targets for zero-shot stance detection;  (iv) \textbf{Branch-BERT}~\cite{li2022improved} utilizes a TextCNN~\cite{TextCNN} to extract important n-grams features incorporating contextual information in conversation threads. All four usages of PLM are based on bert-base-uncased.

\paragraph{In-context Learning with LLMs}

We also conduct experiments with ChatGPT (gpt-3.5-turbo\footnote{https://platform.openai.com/docs/models/gpt-3-5} and gpt-4\footnote{https://platform.openai.com/docs/models/gpt-4-and-gpt-4-turbo}) and LLaMA (LLama 2-70b\footnote{https://huggingface.co/meta-llama/Llama-2-70b-chat-hf}), which are popular and powerful LLMs. Specifically, we employ in-context learning with one demonstration sample.

\begin{table}
\begin{center}	
\resizebox{\linewidth}{!}{
    \begin{tabular}{l|cccccc}
    \hline
    \textbf{Methods}& Bitcoin& Tesla& SpaceX& Biden& Trump & Avg.\\  
    \hline
    \multicolumn{7}{c}{Only considering individual posts/comments}\\ 
    \hline
    
    
        BiLSTM        & 32.99   & 31.40 & 22.79& 25.54 & 24.47 &27.44\\
        TAN            & 33.68 &  33.19   & 25.86 & 26.43 & 25.84& 29.00\\
        GCAE       & 46.25  & 36.70   & 38.37 & 25.42 & 35.34 &36.42\\
        CrossNet       & 32.73  & 31.76   & 30.12 & 20.28 & 30.27&29.03 \\
        
    \hdashline
        MPT       & 49.45  & 41.80  & 46.38 & 27.98 & 36.50 & 40.42\\
        KPT            & 50.34 &  43.11   & 47.47 & 28.90 & 41.87 &42.34\\
        KEPrompt            & 50.34  & 41.23 & 47.11 & 30.31 & 40.87&41.97 \\ 
    \hdashline
        Bert       & 50.99   & 43.72 & 45.88 & 26.65 & 42.45&41.94 \\
        TTS            & 50.88  & 43.85 & 47.50 & 29.00 & 42.10 &42.67\\ 
        JoinCL            & 50.21  & 31.06 & 51.47 & 26.32 & 34.54 &38.72\\
    
    \hline
     \multicolumn{7}{c}{Considering conversation history}\\ 
    \hline
    
    
     BiLSTM        
    & 44.27 
    & 35.55 
    & 28.15   
    & 27.36 
    & 26.47 
    & 32.36
    \\
     TAN            
    & 40.78 
    & 39.31 
    & 28.15   
    & 28.35 
    & 29.31
    & 33.18
    \\
     GCAE       
    & 48.75 
    & 42.75 
    & 42.07  
    & 30.10  
    & 39.43
    & 40.62
    \\
    CrossNet       
    & 37.73 
    & 31.76 
    & 33.63  
    & 25.49 
    & 37.94
    & 33.31
    \\
    \hdashline
         MPT       
    & 51.42 
    & 44.53 
    & 51.30   
    & 31.08 
    & 38.84
    & 43.43
    \\
         KPT            
    & 53.22 
    & 46.67
    & 52.65   
    & 32.22
    & 43.97
    & 45.75
    \\
   
     KEPrompt            
    & 53.22 
    & 45.64  
    & 50.91   
    & 31.08
    & 43.64
    & 44.90
    \\
     \hdashline
         BERT       
    & 53.60 
    & 47.39 
    & 49.31   
    & 29.13 
    & \underline{45.11}
    & 44.91
    \\
     TTS            
    & \underline{53.60} 
    & 46.08  
    & 52.41   
    & 31.23 
    & 44.41
    & 45.55\\
     JoinCL            
    & 52.57 
    & 31.42  
    & 55.03
    & 29.58 
    & 35.04
    & 40.73
    \\
     Branch-BERT         
    & 49.17 
    & 37.14  
    & 37.97   
    & 27.73 
    & 43.07
    & 39.02
    \\
    \hdashline
    LLama 2-70b        
    & 49.88 
    & 46.46  
    & 43.15   
    & \underline{39.17}
    & 36.18
    &42.97
    \\
    gpt-3.5-turbo
    & 46.89
    & \underline{51.69}
    & 53.16  
    & 36.05 
    & 27.47
    & 43.05
    \\
    gpt-4
    &49.39
    &50.71
    &\underline{55.34}
    &\textbf{45.09}
    &40.33
    &\underline{48.17}
    \\
    \hdashline
     \textbf{GLAN}            & \textbf{56.95} & \textbf{52.38}  & \textbf{55.98}   &  38.15 & \textbf{48.91} & \textbf{50.47}\\
     \hline
    \end{tabular}
    }
\end{center}	
\captionsetup{font=small}
\caption{\label{tab:result} Performance of baseline models for in-target stance detection on the five targets in the MT-CSD dataset, considering two experimental settings: ``\textit{Only considering individual posts/comments}'' and ``\textit{Considering conversation history}''.}
\end{table}

\begin{table*}
\begin{center}	
\resizebox{\linewidth}{!}{
\begin{tabular}{c|ccc|ccc|ccc|ccc|ccc}
\hline
\textbf{Target} & \multicolumn{3}{c|}{Bitcoin} & \multicolumn{3}{c|}{Tesla} & \multicolumn{3}{c|}{SpaceX} & \multicolumn{3}{c|}{Biden} & \multicolumn{3}{c}{Trump} \\  
\hline
\textbf{depth} & 1-2 & 3-5  & 6-8 & 1-2 & 3-5  & 6-8& 1-2 & 3-5  & 6-8& 1-2 & 3-5  & 6-8& 1-2 & 3-5  & 6-8  \\ 
\hline
    CrossNet            & 45.98 & 36.23  &35.33 & 41.21 & 38.56  &  31.8& 35.23 & 35.87  & 25.96& 22.86 & 23.18  & 18.65& 24.53 & 37.83  & 37.87  \\
  \hdashline
    KEPrompt          & 54.57 & 55.31 & 38.84 & 28.57 &43.43 & 46.79& 49.41 & 55.67  & 35.53& 34.21 & 30.4  & 29.86& 31.82 & 41.5  & 39.58 \\
\hdashline
    BERT        & 52.14 & 54.79 & 51.74 & 33.33 & 47.24  & 49.02& 47.22 & 50.72  & 37.85& 28.49 & 31.68  & 28.05& 32.2 & 47.56 & 42.77 \\
    TTS            & \textbf{57.08} & 51.92 & 51.63 & \textbf{51.32} & 45.79  & 40.94& 50.38 & 54.16  &48.96& 31.43 &31.54  & 29.94& 31.82 & 48.14  &45.05  \\
    Branch-BERT       & 56.85 &49.23  & 49.5 & 23.81 & 30.76  & 40.86& 41.67 & 39.14 & 33.28& 24.14 & 31.15  & 24.04& 31.67 & 41.97  & 40.76\\ 
\hdashline
    LLama 2-70b          &50.41 & 52.4 &44.59 &35.28& 46.56&47.12 & 42.11&51.03&46.6&39.57&\textbf{38.61}&40.22&\textbf{35.49}&37.17&34.57 \\
    gpt-3.5-turbo          &48.87 &47.38&38.95 &27.75&\textbf{50.6}&53.76&41.65&54.3&53.13&\textbf{46.28} &35.85&33.61&24.73&26.71&26.4\\
\hdashline
    \textbf{GLAN}            & 56.46 & \textbf{59.76}  & \textbf{53.99}& 24.44& 49.42  & \textbf{54.92}& \textbf{50.77} & \textbf{56.95}  & \textbf{53.23}& 28.95 &36.46  & \textbf{42.01}& 33.33 & \textbf{50.25}  & \textbf{47.35}  \\
\hline
\end{tabular}

}
\end{center}	
\caption{\label{tab:depth} Results of different models for the instances with depths 1-2, 3-5, and 6-8 in the setting of considering conversation history.}
\end{table*}

\section{Experimental Results}
In this section, we perform comprehensive experiments on our MT-CSD dataset. Concretely, we present model comparisons in both in-target and cross-target setups. Notably, the reported results are averages obtained from three distinct initial runs.


\subsection{In-Target Stance Detection}
We first report the experimental results on the MT-CSD dataset in the in-target setup, the training and testing sets share identical targets. Two distinct settings are considered in the experiments, involving the utilization of individual posts or comments as input and the consideration of both the current comment and the entire conversation history. The results of these experiments are illustrated in Table~\ref{tab:result}. From the results, we have the following observations. First, the models considering conversations as input consistently outperform their counterparts that take individual sentences as input. This observation underscores the advantages of analyzing stances within the context of conversations. 

\begin{table}
\resizebox{\linewidth}{!}{
    \begin{tabular}{c|ccccc}
        \hline
        \textbf{Target} & CrossNet &  KEPrompt & BERT & TTS  & \textbf{GLAN} \\
        \hline
        \multicolumn{6}{c}{within the same domain} \\ 
        \hline
        DT $\rightarrow$ JB &14.33 &11.75 &20.34 &28.87& \textbf{30.10} \\
        JB $\rightarrow$ DT &15.35 &13.19 &24.87 &30.41& \textbf{31.56} \\
        SX $\rightarrow$ TS &20.09 &20.58 &30.06 &38.78& \textbf{40.08} \\
        TS $\rightarrow$ SX &17.90 & 31.85 & 37.32& 40.06& \textbf{40.85} \\
        \hline
        \multicolumn{6}{c}{across dissimilar domains} \\ 
        \hline
        BC $\rightarrow$ DT & 23.47& 30.23&32.45 & \textbf{32.97}&30.12 \\
        BC $\rightarrow$ JB & 21.29& 29.14&28.34 & \textbf{32.70}&28.78 \\
        BC $\rightarrow$ SX & 26.04& 43.72&40.26 &39.38& \textbf{40.56} \\
        BC $\rightarrow$ TS & 23.73& 36.68&34.84 &36.37& \textbf{38.49} \\
        DT $\rightarrow$ BC &23.94 &15.67 &33.21 &\textbf{35.29}&29.39 \\
        TS $\rightarrow$ BC &21.67 &24.89 &27.65 &\textbf{37.35} &30.18 \\
        SX $\rightarrow$ DT &12.46 &23.18 &36.08 & \textbf{39.58}&32.40 \\
        DT $\rightarrow$ SX &11.88 &23.39 &22.89 &26.27 & \textbf{27.73} \\
        \hline
    \end{tabular}
  }
\caption{\label{tab:cross-target} Comparison of different models for cross-target stance detection.}
\end{table}

Secondly, the performance of LLM methods has been found unsatisfactory, with LLaMA achieving only 42.97\%, while GPT-3.5 Turbo and GPT-4 scored 43.05\% and 48.17\%, respectively, in evaluations across all targets.
This phenomenon could be attributed to the limitations of large models, as their knowledge bases are typically built on historical data and may not accurately capture new targets or events. Third,  GLAN  outperforms almost all baseline models on the MT-CSD dataset. The significance tests comparing GLAN to Branch-BERT, JoinCL, and TTS reveal that GLAN exhibits a statistically significant improvement across most evaluation metrics (with a p-value of $<$ 0.05). Fourth, even state-of-the-art stance detection methods, exemplified by GLAN, exhibit an accuracy of only 50.47\%, highlighting the persistent challenges in conversational stance detection.

\subsection{Cross-Target Stance Detection}
We undertook a series of cross-target experiments on the MT-CSD dataset. The stance detection models are initially trained and validated on a source target and subsequently tested on a destination target. Our experimental design encompasses all available targets, including ``\textit{Bitcoin}'' (BC), ``\textit{SpaceX}'' (SX), ``\textit{Tesla}'' (TS), ``\textit{Joe Biden}'' (JB), and ``\textit{Donald Trump}'' (DT).  Given the dataset's comprehensive coverage across three distinct domains, namely, \textit{cryptocurrency} (BC), \textit{business} (SX, TS), and \textit{politics} (JB, DT), we devise cross-target stance detection experiments, evaluating models both within the same domain and across dissimilar domains. As shown in Table~\ref{tab:cross-target}, our GLAN model exhibits superior performance when training and testing targets are from the same domain when compared to other models. In cross-target experiments across different domains, TTS demonstrates better performance. This observation can be attributed to the similarity of topics within the same domain.

\subsection{Impact of Conversation Depth}
The objective of this analysis is to scrutinize the performance of diverse stance detection models across various conversation depths. The results with different conversation depths are reported in Table~\ref{tab:depth}. Remarkably, our GLAN model consistently achieves the most favorable results for the instances with the depths 6-8. LLMs exhibit excellent performance for the instances with depths 1-2, while they perform much worse than GLAN for the instances with depths 6-8. 

\begin{table}
  \resizebox{\linewidth}{!}{
  \begin{tabular}{lccccl}
    \hline
    Methods & Bitcoin & Tesla & SpaceX & Biden & Trump \\
    \hline
    \textbf{w/o} Global & 48.14 & 50.18 & 49.49& 28.26 & 39.13\\
    \textbf{w/o} Local & 45.95& 49.97 & 48.17& 28.53 & 43.36\\
    \textbf{w/o} Structural  & 48.45& 47.35 & 53.14& 34.64 & 44.00 \\
    \textbf{w/o} Target-attention  & 44.53& 47.02 & 49.73& 27.01 & 45.06 \\
    \hdashline
    \textbf{GLAN}& \textbf{56.95}&\textbf{ 52.38} &\textbf{ 55.98} & \textbf{38.15} & \textbf{48.91}\\
    \hline
  \end{tabular}
  }
\caption{\label{tab:Ablation study} Ablation test results.}
\end{table}


\subsection{Ablation study}
To investigate the influence of different components on the performance of GLAN, we conduct an ablation test of GLAN. This involves removing specific components, including the Global Layer (denoted as w/o Global), which renders the structure akin to conventional attention-based methods, the Local Layer (denoted as w/o Local), the Structural Layer (denoted as w/o Structural), and the Target-attention Layer (denoted as w/o Target-attention). The results of this ablation study for the proposed GLAN are presented in Table ~\ref{tab:Ablation study}.
From the results, we can observe that all the four components have large impact on the performance of GLAN.


\section{Conclusion}
This paper presents MT-CSD, an extensive English conversational stance detection benchmark designed with a specific emphasis on conversation depth. MT-CSD addresses critical challenges in the conversational stance detection task, striving to bridge the gap between research and real-world applications. We devise a GLAN model to address both long and short-range dependencies inherent in conversations. We conduct extensive experiments on our MT-CSD dataset, and experimental results demonstrate that GLAN achieves superior results on the MT-CSD dataset. In addition, extensive experimental findings underscore that MT-CSD poses a more formidable challenge compared to existing benchmarks, as even the state-of-the-art stance detection methods, exemplified by GLAN, achieve an accuracy of only 50.47\%. This highlights substantial opportunities for advancements and innovations in conversational stance detection. In the future, we plan to combine linguistic knowledge and LLMs to further improve the performance of conversational stance detection. 


\section{Acknowledgements}

Min Yang was supported by National Key Research and Development Program of China (2022YFF0902100), Shenzhen Basic Research Foundation (JCYJ20210324115614039). 
Bowen Zhang was supported by National Nature Science Foundation of China (No.62306184), Natural Science Foundation of Top Talent of SZTU (grant no. GDRC202320).
Xiaojiang Peng was supported by the National Natural Science Foundation of China (62176165), the Stable Support Projects for Shenzhen Higher Education Institutions (20220718110918001), the Natural Science Foundation of Top Talent of SZTU(GDRC202131).

\section*{Bibliographical References}\label{sec:reference}

\bibliographystyle{lrec-coling2024-natbib}
\bibliography{lrec-coling2024-example}


\end{document}